# Self-Configuration in Machine Learning

Eugene Wong
University of California at Berkeley

**Abstract**

In this paper we first present a class of algorithms for training multi-level neural networks with a quadratic cost function one layer at a time starting from the input layer. The algorithm is based on the fact that for any layer to be trained, the effect of a direct connection to an optimized linear output layer can be computed without the connection being made. Thus, starting from the input layer, we can train each layer in succession in isolation from the other layers. Once trained, the weights are kept fixed and the outputs of the trained layer then serve as the inputs to the next layer to be trained. The result is a very fast algorithm.

The simplicity of this training arrangement allows the activation function and step size in weight adjustment to be adaptive and self-adjusting. Furthermore, the stability of the training process allows relatively large steps to be taken and thereby achieving in even greater speeds. Finally, in our context configuring the network means determining the number of outputs for each layer. By decomposing the overall cost function into separate components related to *approximation* and *estimation*, we obtain an optimization formula for determining the number of outputs for each layer.

With the ability to self-configure and set parameters, we now have more than a fast training algorithm, but the ability to build automatically a fully trained deep neural network starting with nothing more than data.

**Introduction**

Deep neural networks (DNN) are those with multiple layers (say, >2) and "deep learning" refers to procedures for adjusting the weights so that the final output of the DNN matches the targeted output. Back propagation is a standard deep learning algorithm that is mathematically elegant, but difficult to use. Nonetheless, with skillful modifications and tuning, deep learning with back propagation has met spectacular success. In applications involving tasks normally undertaken by human beings, such as speech recognition and medical diagnosis, properly trained DNNs have often exceeded human performance.

As Bengio et. al. note (Bengio 2006) failures tend to go unreported so that the real difficulty of using back propagation, though recognized, is probably substantially underreported and underrated. Training a DNN *one layer at a time* is clearly an attractive option, and a substantial literature now exists on this subject.

In training a neural network, one starts with nothing more than a set of input vectors and, for each input, a target output vector. A subset of the (input, target) pairs is used to train the network, and the remainder for testing. Clearly, training one layer at a time has to start with the input layer. But measuring how well the final output matches the target output is several layers away. Thus, most of the existing papers on "one layer at a time" deal with unsupervised learning (Bengo 2006, Hinton 2007, Lengelle 1996) where the weight adjustment is done according to criteria that do not directly involve the error at the output. Instead, they are chosen to extract features or to make the data in the training set better behaved in some way. A major exception is the recent work of Hettinger et. al. (Hettinger 2017). Their procedure is clearly supervised learning, but not truly one layer at a time. It is best characterized as "train two, keep one." For each layer to be trained, one connects it to an output layer and uses a supervised algorithm, such as back propagation, to train both layers. Upon completion the output layer is then discarded.

In this paper we present an algorithm that is supervised and truly one layer at a time. Training is done using no information beyond the layer being trained and the weights are adjusted in a direct effort to reduce the final error. We believe that our algorithm has superior performance is several ways. These include fast training speed and good final training solution.

Additional advantages include the fact that the operating parameters of each layer are easy to set and design features such as the number of weights can be decided at training time. These characteristics make it possible to have self-configuring, self-tuning, and self-training neural networks that are particularly well suited for use in autonomous systems such as robots and self-driving vehicles. The same issues are dealt with in (Chatterjee 2017) but the approach there is highly computation-intensive and not suitable for the autonomous systems which are our primary focus.

**Neural Network**

In this paper we only consider fully connected neural networks, one layer of which is shown in Figure 1. Extension to other trainable machines, such as convolutional neural networks, is clearly possible.

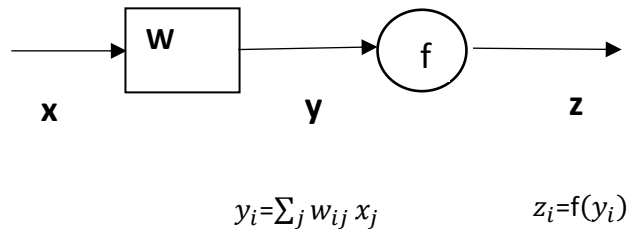

$y_i = \sum_j w_{ij} x_j$ $\qquad$ $z_i = f(y_i)$

Figure 1. One Layer of Neural Network

The nonlinear function f is known as the *activation function* and can be any increasing function that satisfies the condition:

$\qquad$ $f(y) \cong ay$ $\qquad$ for $|y| \cong 0$ $\hfill$ (1)
$\qquad\qquad\quad \cong b\, \text{sign}(y)$ as $|y| \to \infty$

A common choice is the symmetric sigmoid function

$$f(y) = b\left(\frac{1-e^{-\alpha y}}{1+e^{-\alpha y}}\right) \quad \text{with } \alpha = \frac{2a}{b} \tag{2}$$

However, our preference is for the *rectifying amplifier* function

$$\begin{aligned} f(y) &= ay & \text{for } a|y| \le b \\ &= b\,\text{sign}(y) & \text{otherwise} \end{aligned} \tag{3}$$

A neural network is made up of a number of layers connected in series so that the output of layer m is the input of layer m+1 as shown in Figure 2.

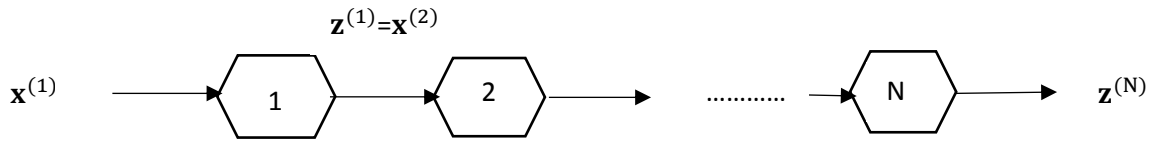

Figure 2. Neural Network

We shall also denote the input to the overall neural network as **x** and the final output as **z**. Thus,

**x**= $\mathbf{x}^{(1)}$ and **z**=$\mathbf{z}^{(N)}$.

**Supervised Training**

For training a neural network, we are given a set of sample input vectors **x** and for each input a target output **t(x).** The task of training is to adjust the weights of every layer so that for every sample in the training set the error **z(x)-t(x)** is as small as possible. This is usually done by minimizing a cost function C of the form

$$C = \sum_{\mathbf{x}} h(\mathbf{z}(\mathbf{x}), \mathbf{t}(\mathbf{x})) \tag{4}$$

Our choice is the quadratic function

$$C = (1/2) \sum_{\mathbf{x}} \|\mathbf{z}(\mathbf{x}) - \mathbf{t}(\mathbf{x})\|^2 \tag{5}$$

Often a penalty term dependent only on the weights is added to the cost in order to limit excessive growth in the weights during training. Since we train only one layer at a time that is not likely to be needed. In any event it would only require a small modification.

**Back Propagation**

Back propagation can be viewed as a neural-like network propagating the error signal $\mathbf{z}(\mathbf{x}) - \mathbf{t}(\mathbf{x})$ from the output layer (the Nth) of the neural network back to the input layer. A typical layer of the backward network is depicted in Figure 3.

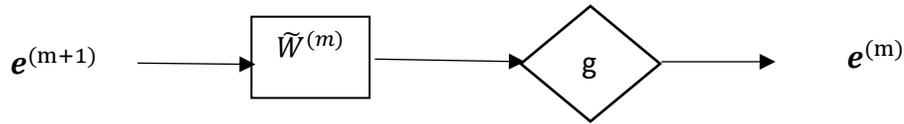

Figure 3. Error Propagation

where $\widetilde{W}$ denotes transpose of $W$ and g multiplies the ith coordinate by $f'\left(y_i^{(m)}\right)$. The precise formula for the error signal is given by

$$e_i^{(m)} = f'\left(y_i^{(m)}\right) \sum_j w_{ji}^{(m)} e_j^{(m+1)} \tag{6}$$

where f' denote differentiation and we set $e_i^{(N+1)} = f'(y_i^{(N)})\,(z_i - t_i)$ .

**Gradient Descent**

We note that if we change the weights incrementally by

$$\Delta w_{ij}^{(m)} = -\delta \frac{\partial C}{\partial w_{ij}^{(m)}} \qquad \delta > 0 \tag{7}$$

then the change in C is approximately given by

$$\Delta C \cong -\delta \sum_{m,i,j} \left(\frac{\partial C}{\partial w_{ij}^{(m)}}\right)^2 \leq 0 \tag{8}$$

This is known as Gradient descent. We use this method of changing weights throughout this paper. For the quadratic cost function, we can write

$$\frac{\partial C}{\partial w_{ij}^{(m)}} = \sum_x e_i^{(m)} x_j^{(m)} \tag{9}$$

This provides us with a weight change formula

$$\Delta w_{ij}^{(m)} = -\delta \sum_x e_i^{(m)} x_j^{(m)} \qquad (10)$$

and completes the back propagation algorithm for adjusting the weights.

**Training One Layer at a Time**

Choosing a linear output layer is common throughout the history of neural network. Indeed the nonlinear layers were introduced initially only to increase the dimension of the input space to make the samples more linearly separable. Thus, the widely used Support Vector Machines (SVM) and Radial Basis Function Machines (RBFM) all have linear output layers. (Both SVM and RBFM have good descriptions on Wikipedia.) In 2004, a class of neural networks with linear output layers called Extreme Learning Machines was introduced (Huang 2004). These have attracted a large following, but also severe criticisms for re-invention without adequate attribution (See e.g., the web page: elmorigin.wixsite.com/originofelm).

If we remove the nonlinearity from the output layer, then for a quadratic cost function the optimal weights have a closed form expression in terms of the input to the output layer and the target **t**. We now propose to train a deep neural network one layer at a time by using the effect of being connected to an optimized linear output layer without the actual connection being made. The arrangement is shown in Figure 4.

At this point it is useful to represent the entire training set as a matrix **X** whose columns are the sample input vectors arranged in an arbitrary but fixed order. All the other vectors in the neural network are arranged accordingly as matrices as well and denoted by bold upper case letters. For example, the output of the neural network is now **Z**, error signal **E** and the target **T**. With this representation, we can eliminate summations over **x** in all linear operations.

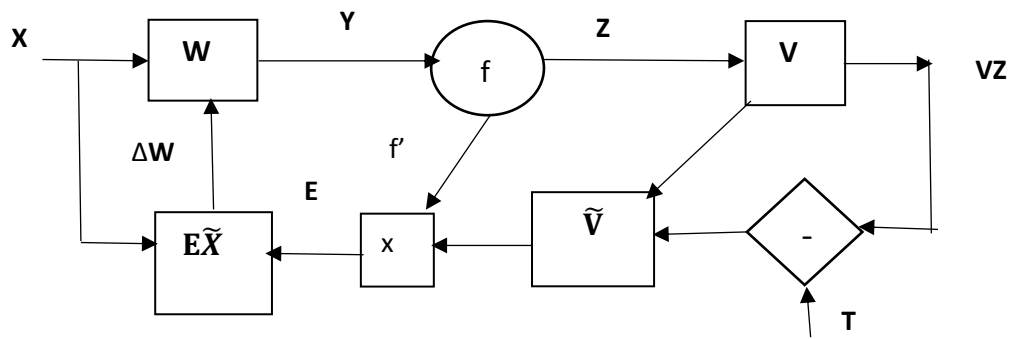

Figure 4. Weight Adjustment Loop

The weights in the optimal linear output layer are given by the matrix **V** which satisfies the orthogonality condition: **(VZ-T)Z̃** = 0, or

$$\mathbf{VA} = \mathbf{T\tilde{Z}} \tag{11}$$

where $\mathbf{A} = \mathbf{Z\tilde{Z}}$ is a semi-positive definite symmetric matrix. One solution of (11) for **V** is

$$\mathbf{V} = \mathbf{TZ}^+ \tag{12}$$

where $\mathbf{Z}^+$ denotes the Moore-Penrose inverse of **Z**. If **A** is invertible, which is nearly always the case, then $\mathbf{Z}^+ = \mathbf{\tilde{Z}A}^{-1}$ and **V** is unique. The weight matrix in the output layer is then given by

$$\mathbf{V} = \mathbf{T\tilde{Z}A}^{-1} = \mathbf{T\tilde{Z}}(\mathbf{Z\tilde{Z}})^{-1} \tag{13}$$

which shows that **V**, and thus also Δ**W**, depend only on **T** and **Z**., and we have

$$\Delta\mathbf{W} = -\delta\ \mathbf{E\tilde{X}} \tag{14}$$

Since **T** remains fixed, each cycle of training depends only on **Z** for that cycle. To be sure, **V** has to be recomputed for each cycle, but is done separately. Even when **A** is singular, **V** can be computed quickly, by a separate training cycle if necessary.

Once the training for a layer is complete, the weights are kept fixed and the final output **Z** is preserved and used as input to the next layer to be trained.

**Setting the Parameters**

There are three parameters associated with the layer to be trained: the constants a and b in equation (1) for the activation function, and the step size constant δ for the weight change in equations (7), (8), (10) and (14). In actual fact these are only *constant* during each cycle of training. Before dealing with setting these parameters for each cycle, we make one minor modification. Let μ denote the mean value of the components of the matrix **Y**. We define $z_{ij} = f(y_{ij} - \mu)$ rather than $z_{ij} = f(y_{ij})$. Centering the components of **Z** avoids having to deal with biases in the weights.

Assume that we initialize the training by setting the weights in **W** as random numbers between -1 and 1. The objective in setting the three parameters is to keep the weights roughly in the same range as training proceeds, and to keep the components of **Z** in the same range. The scaling control cannot be exact, but has to be reexamined as more empirical experience is accumulated. To achieve these objectives, we set the parameters a and b in the activation as follows:

$$a = \left(\sum_{y \in \mathbf{Y}}(y - \mu)^2\right)^{-.5} \tag{15}$$

And choose b to be the largest number such that

$$\text{Fraction}(\ y \in \mathbf{Y} : \text{abs}(y) > b) > .3 \tag{16}$$

Such a value for b always exists if a is chosen according to (15).

To choose the constant δ in (14), we set

$$\delta = .15\left(\sum_{u \in \mathbf{E}\tilde{X}} u^2\right)^{-.5} \qquad (17)$$

which is intended to keep the values of ΔW about .15 of the values of **W.**

The number .15 is determined empirically and is rather aggressive (i.e., large step size). Typically, if one observes the cost function evaluated on the test data during training, the plot would be somewhat jagged, indicating a degree of oscillation. We note that oscillatory behavior during training is not a problem since using the "best up to now" results is itself a legitimate algorithm provided we keep the weight information for the "best up to now" results. A plot for the "best up to now" results for the cost function is always smooth. Thus, setting step size large to achieve a faster descent is almost always a good trade off unless there is instability in the algorithm. We have not seen such instability in the examples that we have tested.

**Configuring the Network**

At the beginning of a DNN machine learning application, all we are given is a collection of sample vector-pairs (**x, t**). In our one-layer-at-a-time context, we adopt the following approach to configuring the network:

a. The network will always have a nonlinear input layer and a linear output layer.
b. A layer will be added if it improves the cost function on the test data.
c. The process of configuring the network stops when the next candidate layer is not added.

In this process the number of inputs to each layer is always known, so the entire configuration problem is reduced to one of deciding on the number of outputs (i.e., the dimension of the output vector) for the next layer to be added. The problem is recursive so that we only need to deal with the problem for the input layer.

We begin by noting that a reasonably general formulation of the machine learning problem is as follows: Let $f: R^n \to R^n$ be an unknown function. We want to find an approximation $\hat{f}$ to $f$ in the following way:

First, choose a parametric family of functions $\hat{f}_\mathbf{w}$, where $\mathbf{w} \in R^k$ is a set of parameters that are called weights in the case of neural networks and chosen to make $\hat{f}_\mathbf{w}$ a good approximation to $f$.

Second, use a collection of sample pairs $\mathbf{S} = \{(x, f(x))\}$ to obtain an estimate $\hat{\mathbf{w}}$ of the parameters **w.**

This is exactly the situation in machine learning for neural networks. There if the samples are from a pair of random variables (**x, t**) and the cost function is quadratic then the result will be an approximation to the conditional expectation: $f(\mathbf{x}) = E(\mathbf{t}|\mathbf{x})$.

Let **x** be a random variable. Assume that the weights **w** are to be estimated using a collection of independent samples. There are two unrelated sources of error in the approximation:

    a. Approximation error $\qquad f(\mathbf{x}) - \hat{f}_\mathbf{w}(\mathbf{x})$

    b. Estimation error $\qquad \hat{f}_\mathbf{w}(\mathbf{x}) - \hat{f}_{\hat{\mathbf{w}}}(\mathbf{x})$

Because these errors are unrelated, we assume that we can write

$$E[f(\mathbf{x}) - \hat{f}_{\hat{\mathbf{w}}}(\mathbf{x})]^2 = E[f(\mathbf{x}) - \hat{f}_\mathbf{w}(\mathbf{x})]^2 + E[\hat{f}_\mathbf{w}(\mathbf{x}) - \hat{f}_{\hat{\mathbf{w}}}(\mathbf{x})]^2 \qquad (18)$$

Let N denote the number of sample pairs in **S**. We now want to examine the relationship between and the integers: *n, m, k,* and *N* and the two terms on the right side of (18).

Consider a two-layer arrangement with a nonlinear input layer connected to a linear output layer. There are *n* inputs and *m* outputs. Let *p* be the number of outputs to the input layer. Then *k*, the number of weights in this two layer arrangement is given by

$$k = (n + m)p \qquad (19)$$

The first step in configuring the input layer is to determine an appropriate number for *p*.

We note that the approximation error depends only on *k*. If the approximation is a series expansion, then typically the approximation error is proportional to $k^{-\lambda}$ where λ is some positive constant. The sampling error, in turn, should be proportional to *k*, one error per weight, and inversely proportional to the number of samples *N* (noise variance decrease as 1/*N*). Hence, from (18) we can write

$$\sigma^2(k) = E[f(\mathbf{x}) - \hat{f}_{\hat{\mathbf{w}}}(\mathbf{x})]^2 = \alpha k^{-\lambda} + \beta\left(\frac{k}{N}\right) \qquad (20)$$

where α, β and λ are positive constants.

Although it might seem that we have merely traded one set of unknown parameters for another such set, the constants α, β and λ can actually be determined empirically. Consider the two-layer arrangement with $k = (n + m)p$ and assume $k \ll N$. Then the last term can be omitted in an approximation and we have

$$\ln(\sigma^2(k)) \cong \ln(\alpha) - \lambda \ln(k) \qquad (21)$$

Then α and λ can be determine by measuring $\sigma^2(k)$ for small values of $k$ and using linear regression. The remaining constant β can then be determined by measuring $\sigma^2(k)$ for a larger value of $k$.

If we ignore the constraint that $k$ must be an integer for the moment, we find the value of $k$ that minimizes $\sigma^2(k)$ to be

$$k_o = (\frac{\alpha\lambda}{\beta})N^{(1/(\lambda+1))} \tag{22}$$

from which $p_0$, the optimum number of outputs to the input layer, can now be determined.

To use our method for computing $p_0$ for the input layer, we need to train a few small configurations (say, p = 1 to 4) in order to compute α and λ and one of moderate size (say p = 8) to compute β. The values for α and λ need not be changed for successive layers since the approximating structure remains the same. But the parameter β has to be recomputed for each layer since the input samples are different for each layer.

A very quick, but crude, calculation for $p_0$ can be made by applying the methodology to *untrained* configurations, say those with random weights in the matrix **W**. But one must remember that now the only adjustable weights are in **V**, hence $k = mp$.

Limited experimentation has provided excellent support to the model given by (20) and the approach of data driven configuration that we propose.

**Conclusion**

The approach of one layer at a time provides us with more than a fast algorithm, but, more importantly, the ability to build a fully trained deep neural network starting with nothing more than data.

**Acknowledgement**

I want to thank Professor Bruce Hajek of the University of Illinois, Urbana-Champaign, for a careful reading of the manuscript and providing many suggestions for improvement. I want to thank Professor C. Y. Lee and Eugene Lee at the National Chiao-Tung University in Taiwan for experimental testing of the algorithms, and Dr. Jan Ming Ho of Academia Sinica for his interest and support.

**References**

Bengio, Y., Lamblin, P., Popovici, D., and Larochelle, H. (2006). "Greedy Layer-Wise Training of Deep Networks," *NIPS'06 Proceedings of the 19th International Conference on Neural Information Processing Systems,* 153-160, December 04-07, 2006.

Chatterjee, s., Javid, A. M., Sadeghi, M., Mitra, P. P. and Skoglund, M. (2017) "Progressive Learning for Systematic Design of Large Neural Networks," *arXiv*: 1710.08177v1 [cs.NE] Oct. 23, 2017.
Hettinger, C., Christensen, T., Ehlert, B., Humpherys, J., Jarvis, T. and Wade, S., "Forward Thinking: Building and Training Neural Networks One Layer at a Time," *arXiv*:1706.02480.v1 [stat.ML], June 8, 2017.

Hinton, G. E., "Learning Multiple Layers of Representation,"
*Trends in Cognitive Sciences* 11, 428-34, 2007.

Huang, G. B. and Siew, C. K., "Extreme Learning Machine: RBF Network Case," *Proc. Int. Conf on Control, Automation, Robotics and Vision (ICARCV)* 1029–1036, 2004.

Lengelle, R. and Denceaux, T., "Training MLP's Layer by Layer Using an Objective Function for Internal Representations," *Neural Networks* 9, 83-97, 1996.

(U.S. patent pending)